\def\isarxiv{1} 

\ifdefined\isarxiv
\documentclass[11pt]{article}

\usepackage[numbers]{natbib}

\usepackage{subfig}
\usepackage{wrapfig,epsfig}

\else
\documentclass{article}
\usepackage{neurips_2023}

\usepackage{natbib} 

\fi

\usepackage{amsmath}
\usepackage{amsthm}
\usepackage{amssymb}
\usepackage{algorithm}
\usepackage{algpseudocode}
\usepackage{graphicx}
\usepackage{grffile}
\usepackage{url}
\usepackage{xcolor}
\usepackage{epstopdf}
\usepackage{hyperref}           

\usepackage{bbm}
\usepackage{dsfont}

\allowdisplaybreaks

\ifdefined\isarxiv

\usepackage{tikz}
\usepackage{hyperref}  
\hypersetup{colorlinks=true,citecolor=blue,linkcolor=blue} 
\usetikzlibrary{arrows}
\usepackage[margin=1in]{geometry}

\else

\usepackage{microtype}
\usepackage{hyperref}
\definecolor{mydarkblue}{rgb}{0,0.08,0.45}
\hypersetup{colorlinks=true, citecolor=mydarkblue,linkcolor=mydarkblue}

\fi

\newtheorem{theorem}{Theorem}[section]
\newtheorem{lemma}[theorem]{Lemma}
\newtheorem{definition}[theorem]{Definition}

\newtheorem{corollary}[theorem]{Corollary}

\newtheorem{remark}[theorem]{Remark}
\newtheorem{claim}[theorem]{Claim}

\newcommand{\wh}{\widehat}
\newcommand{\wt}{\widetilde}

\newcommand{\R}{\mathbb{R}}

\renewcommand{\hat}{\wh}
\renewcommand{\k}{\mathsf{K}}

\DeclareMathOperator*{\E}{{\mathbb{E}}}

\DeclareMathOperator{\poly}{poly}

\DeclareMathOperator{\nnz}{nnz}

\makeatletter
\newcommand*{\RN}[1]{\expandafter\@slowromancap\romannumeral #1@}
\makeatother

\newcommand{\lianke}[1]{{\color{orange}[Lianke: #1]}} 

\usepackage{lineno}

\begin{document}

\ifdefined\isarxiv

\date{}

\title{Efficient SGD Neural Network Training via Sublinear Activated Neuron Identification} 
\author{
Lianke Qin\thanks{\texttt{lianke@ucsb.edu}. UCSB. }
\and 
Zhao Song\thanks{\texttt{zsong@adobe.com}. Adobe Research.}
\and 
Yuanyuan Yang\thanks{\texttt{yyangh@cs.washington.edu}. The University of Washington.}
}

\else

\title{Efficient SGD Neural Network Training via Sublinear Activated Neuron Identification} 
\maketitle

\fi

\ifdefined\isarxiv
\begin{titlepage}
  \maketitle
  \begin{abstract}

Deep learning has been widely used in many fields, but the model training process usually consumes massive computational resources and time. Therefore, designing an efficient neural network training method with a provable convergence guarantee is a fundamental and important research question. In this paper, we present a static half-space report data structure that consists of a fully connected two-layer neural network  
for \emph{shifted ReLU} activation to enable activated neuron identification in sublinear time via geometric search. We also prove that our algorithm can converge in $O(M^2/\epsilon^2)$ time with network size quadratic in the coefficient norm upper bound $M$ and error term $\epsilon$.

  \end{abstract}
  \thispagestyle{empty}
\end{titlepage}

{
}
\newpage

\else

\begin{abstract}

Deep learning has been widely used in many fields, but the model training process usually consumes massive computational resources and time. Therefore, designing an efficient neural network training method with a provable convergence guarantee is a fundamental and important research question. In this paper, we present a static half-space report data structure that consists of a fully connected two-layer neural network  
for \emph{shifted ReLU} activation to enable activated neuron identification in sublinear time via geometric search. We also prove that our algorithm can converge in $O(M^2/\epsilon^2)$ time with network size quadratic in the coefficient norm upper bound $M$ and error term $\epsilon$.

\end{abstract}

\fi

\section{Introduction}


Deep learning is widely used in computer vision~\cite{lbl+98,lhb+99, slj+15,ksh12,hzr+16}, natural language processing~\cite{cwb+11, kjt19}, game playing~\cite{shm+16, sss+17} and beyond. It's often the case that the training of deep learning algorithms takes an enormous amount of computational resources. A fundamental challenge in this line of research is, therefore, designing an efficient neural network training method that provably converges. Existing work that provably converges suffers a over-parameterized network structure~\cite{dzp+18, ll18, adh+19, als19, als19b, dll+19, cg19, sy19, zg19, zmg19, os20, zpd+20, bpsw21, hlsy21}.

The preceding study by~\cite{d20} established that SGD is capable of learning polynomials with restricted weights and particular kernel spaces utilizing a neural network of depth two of \emph{near optimal} size. This is particularly the case for the set of even polynomials of limited degree and with a coefficient vector norm that does not exceed $M$., for input distribution on a unit sphere, $\wt{O}(M^2/\epsilon^2)$ neurons and $O(M^2/\epsilon^2)$ iterations suffice to output a predictor with error $\leq \epsilon$ via depth two neural networks. However, their algorithm still suffers a cost per iteration as $O(mbd)$, where $b$ is the batch size, $m$ is the width of the neural tangent kernel and $d$ is the dimension of input data point. This linear dependency on $m$ comes from computing inner products between the gradient of the loss function with respect to weights and the gradient of the weights with respect to the points being queried. 
This seems to be a natural barrier. One natural question to ask is,
\begin{center}
    {\it Is there some algorithm that only requires $o(mbd)$ cost per iteration?}
\end{center}

We provide a positive response to the preceding question, summarizing our contributions in the following manner:

\begin{itemize}
    \item We proposed a static half-space report data structure that consists of a fully connected two-layer neural network with \emph{neural tangent kernel} for \emph{shifted ReLU} activation. In specific, we build a half-space report data structure of weights with batched SGD update. At every iteration, our algorithm identifies the weights that are fired by current data points and propagates them efficiently. Additionally, we can show that at any given iteration, the number of activated neurons for each input data point is upper bounded by $o(m)$. Thus, via geometric search, our algorithm identifies those activated neurons in time sublinear in $m$.
    \item We show that our algorithm can converge in $O(M^2/\epsilon^2)$ time with network size quadratic in the coefficient norm upper bound $M$ and error term $\epsilon$. 
\end{itemize}

\subsection{Our Results}\label{subsec:result_overview}

To formally introduce our main results, we first present the definitions regarding the input data distribution. In this paper, we will assume that the input data distribution is on a unit sphere $\mathbb{S}^{d-1}$. On top of that, we present the definition of $R$-bounded distribution.

\begin{definition}[$R$-bounded distribution]\label{def:R_bounded_dis}
A distribution ${\cal D}$ on $\mathbb{S}^{d-1}$ is said to be $R$-bounded if, for every $u \in \mathbb{S}^{d-1}$, the expectation $\E_{x \sim {\cal D}} [\langle u,x \rangle^2]$ is less than or equal to $\frac{R^2}{d}$.
\end{definition}

In practice, any given distribution ${\cal D}$ is bounded by $\sqrt{d}$. Additionally, many commonly used distributions are bounded by $O(1)$, or even $(1+o(1))$. Examples of such distributions include uniform distribution on a unit sphere $\mathbb{S}^{d-1}$, on a discrete cube $\{ \pm \frac{1}{\sqrt{d}}\}^d$, and on $\Omega(d)$ randomly selected points.

Then, we present the formal definition of the neural network structure that we will consider in this paper. Our study will focus on fully connected, depth-2 neural networks with $2m$ hidden neurons, implementing a shifted ReLU activation function $\sigma_{b_0} : \R \rightarrow \R$ with $\ell_2$-loss function. More formally, we present the definition of prediction function and loss functions as follows:

\begin{definition}[Prediction and Loss function]\label{def:prediction_loss}
Given $b_0 \in \R, x \in \R^d, W \in \R^{d \times 2m}$ and $a \in \R^{2m}$, we say a prediction function $f$ is $2\mathsf{NN} (2m,b_0)$ if: 
\begin{align*}
    f(W,x,a) := \frac{1}{\sqrt{2m}} \sum_{r=1}^{2m} a_r \sigma_{b_0} (\langle w_r,x \rangle)\\
    l(W) :=\frac{1}{2} \sum_{i=1}^{n} ( f(W,x_i,a)-y_i )^2 
\end{align*}
where $\sigma_{b_0}$ denote the shifted ReLU function $\sigma_{b_0}(x) = \max\{0,x-b_0\}$. 
\end{definition}


The \emph{shifted} ReLU function is frequently employed in literature, as well as theoretical investigations as indicated in~\cite{zpd+20,syz21}. Regarding the neural network weights, we leverage the \emph{Xavier initialization with zero outputs} method as described in~\cite{gb10}. This involves organizing the neurons into pairs, with each pair composed of two neurons initialized identically, differing only by a factor of $\pm$.
We denote ${\cal N}_{d,m}^\sigma := \{h_W(x) = \langle  u, \sigma(W_0 x) \rangle\}$, and $W := (W_0,u)$ as the aggregation of all weights. More formally, we present our weight initialization as follows:

\begin{definition}[Weights at Initialization]\label{def:weight_init}
Given input dimension $d$, number of neurons $2m$, constant $B >0$, we say the weight $W = (W_0,u)$, $W_0 \in \R^{2m \times n}$ is initialized according to distribution ${\cal I} (d,m,B)$ if: 
\begin{itemize}
    \item For each $r \in [m]$, we sample $w_r(0) \sim {\cal N}(0, I_d)$, and $w_{m+r}(0) = w_r(0)$.
    \item For each $r \in [m]$, we sample $a_r$ from $\{+B, -B\}$ uniformly at random, and $a_{m+r} = a_r$.
\end{itemize}
\end{definition}
 Note that if $W \sim {\cal I} (d,m,B)$, then with probability $1$, $h_W(x) = 0,~\forall x$.

 Next, we present a benchmark of our algorithm: 

\begin{definition}[Even Polynomial with Bounded Coefficient Norm]\label{def:poly_bound_norm}
We denote the class of even polynomials with coefficient norm bound $M$ as ${\cal P}_c^M$. More formally,
\begin{align*}
    {\cal P}_c^M := \Big\{ & ~ p(x) = \sum_{|\alpha|~\text{is even and } \leq c} a_{\alpha} x^\alpha 
     : \sum_{|\alpha|~\text{is even and } \leq c} a_\alpha^2 \leq M^2 \Big\}
\end{align*}
\end{definition}

 We are now ready to state our main theorem (a combination of Theorem~\ref{thm:shifted_relu_nnt} and Lemma~\ref{lem:runtime}) that, 
Algorithm~\ref{alg:nnt} is capable of learning even polynomials of bounded norm, denoted ${\cal P}_c^M$, exhibiting near-optimal characteristics in terms of sample complexity and network size. Furthermore, it showcases a per iteration time that is sublinear in $m$:

\begin{theorem}[Main theorem]\label{thm:main_theorem}

Given the following:
\begin{itemize}
    \item a constant $c>0$,  accuracy parameter $\epsilon$, along with positive constants $B > 0$ and $\eta > 0$,
    \item  a selection of parameters $ m = \wt{O}({d^{-1} \epsilon^{-2}M^2 R^2}), T = O({\epsilon^{-2}M^2})$,
    \item sample access to $R$-bounded distribution ${\cal D} \in \R^d$(Definition~\ref{def:R_bounded_dis}),
    \item  input dimension $d$ and coefficient norm bound $M$(Definition~\ref{def:poly_bound_norm}),
    \item ${\cal L}_{\cal D}(h) := \E_{(x,y) \sim {\cal D}} l(h(x),y)$ is the expected loss of predictor $h$ on input distribution ${\cal D}$.
\end{itemize}
there exists an algorithm (Algorithm~\ref{alg:nnt}) which gives that when running the Stochastic Gradient Descent
 \begin{itemize}
     \item  with a batch size of $b$,
     \item  on $2\mathsf{NN}(m,b_0=\sqrt{0.4 \log( 2m)})$(Definition~\ref{def:prediction_loss}),
     \item   returns a function $h$ that satisfies:
    \begin{align*}
        \E[{\cal L}_{\cal D} (h)] \leq {\cal L}_{\cal D} ({\cal P}^M_c) + \epsilon
    \end{align*}
    \item with expected per-iteration running time in
    \begin{align*}
       \wt{O}(m^{1-\Theta(1/d)} b d). 
    \end{align*}
 \end{itemize}

\end{theorem}

\paragraph{Roadmap.} We first present a technique overview of our paper in Section~\ref{sec:tech}. We then introduce some notations and preliminaries in Section~\ref{sec:prelim}. We present our main algorithm and give the proof of correctness of our algorithm in Section~\ref{sec:proof_of_correctness}. We give the proof of the running time of our algorithm in Section~\ref{sec:proof_time}. We conclude the contribution of this paper in Section~\ref{sec:conclusion}.

\section{Technical Overview}\label{sec:tech}

Our work consists of two results: The first result focuses on proving the learnability of \emph{shifted} ReLU activation on two-layered neural network with SGD update. The second result focuses on designing an efficient half-space reporting data structure based on the weight sparsity induced by \emph{shifted ReLU} activation function, which gives us time per iteration sublinear in $m$.

We prove the first result via reduction-based techniques. In order to prove the learnability of shifted ReLU activation on two-layer neural network (Theorem~\ref{thm:shifted_relu_nnt}), we actually prove a more general statement in Theorem~\ref{thm:nnt}, where for general activation function $\sigma$, the expected loss of the two-layer neural network is optimal up to an additive $\epsilon$ error with $m = \wt{O}(M^2/\epsilon^2)$ neurons and $T = O(M^2/\epsilon^2)$ updates. Next, we prove that the general activation on two-layer network is equivalent to neural tangent kernel training (Lemma~\ref{lem:eqa_nn_ntk}) when weight vectors of the network on neurons are initialized with large enough $B$ (Definition~\ref{def:weight_init}). In specific, when $B$ is large enough, SGD with a general activation function on two-layer neural network follows a lazy update, where the weight vector on the input only moves around a small ball. In this case, the first-order approximation of the network function on the initial weight is approximately equivalent to the original network function. In this light, neural tangent kernel learning suffices to approximate the general two-layer neural network training with SGD update. Then, we analyze the neural tangent kernel training in the language of (vector) random feature scheme, which completes the whole proof for the first result. 

The second result is based on the observation that, for \emph{shifted} ReLU activation function on two-layer neural network with SGD update, the number of fired neuron for each data point is sublinear in the network size $2m$. In this light, we adapted half-space reporting data structure \textsc{hsr} to boost the per-iteration running time by preprocessing the network weights. More precisely, the algorithm initializes the half-space reporting data structure \textsc{hsr} with initialized weight vectors, then at every batched SGD iteration, the algorithm queries the weight vectors that fire for the points in the batch by \textsc{hsr} and update the fired neuron set in $\wt{O}(bm^{1-\Theta(1/d)}d)$ time.


\section{Preliminaries}\label{sec:prelim}

\paragraph{Notations}
We use $\sigma$ to denote the descent activation function. We use ${\cal D}$ 
to denote the input distribution. In a matrix $W$, the $i$-th row is represented as $w_i$. We denote the $i$-th row in a matrix $W$ by $w_i$. We use $\|x\|_p = (\sum_{i=1}^{d} |x_i|^p)^\frac{1}{p}$ to represent the $p$-norm of $x \in \R^d$. Given a matrix $W$, we use $|W|$ to denote its spectral norm, defined as $|W| = \max\{ \|Wx\| | \|x\|=1 \}$. We adhere to the standard convention where $|x| = |x|_2$. Regarding a distribution $\cal D$ 
on a space ${\cal X}$, $p \geq 1$ and $f:{\cal X} \rightarrow \R$, we use $\|f\|_{p,{\cal D}} = (\E_{x \sim {\cal D}}|f(x)|^p)^{\frac{1}{p}}$. For any function $f$, we use $\wt O(f)$ to represent $O(f\cdot\poly \log(f))$. For an integer $n$, we use $[n]$ to denote the set $\{1,\ldots, n\}$. For a given function $\sigma$, we leverage $\sigma'$ and $\sigma''$ to represent its first-order and second-order derivative, respectively.

\subsection{Definitions}
In this section, we present some definitions of properties on a function $l$. To begin with, we present the definition of a convex function.

\begin{definition}[Convexity]\label{def:convex}
We say a function $l$ is convex if for any $x_1, x_2$, we have:
\begin{align*}
    f(x_1) \geq f(x_2) + f'(x_2) (x_1 -x_2)
\end{align*}
\end{definition}

Next, we present the definition of Lipschitzness of a function. 

\begin{definition}[$L$-Lipschitz]\label{def:L_lipschitz}
We say a function $l$ is $L$-Lipshitz with respect to norm $\| \cdot \|$ if for any $x_1, x_2$, we have:
\begin{align*}
    |l(x_1) - l(x_2)| \leq L \| x_1 - x_2\|
\end{align*}
\end{definition}

\subsection{Neural Network Training}

Then, we introduce some basic definitions regarding neural network training of supervised learning and some related notations:

\begin{definition}[Supervised Learning]\label{def:supervised_learning}
The objective of supervised learning is to learn a mapping from an input space, denoted as ${\cal X}$, to an output space, denoted as $\cal Y$, using a sample set $S = {(x_i,y_i)}_{i \in [n]}$. These samples are independently and identically drawn from a distribution ${\cal D}$, which spans across ${\cal X} \times {\cal Y}$.
\end{definition}

A special case of the supervised learning is binary classification, where the prediction is a binary label.

\begin{definition}[Binary Classification]\label{def:binary_classification}
The binary classification problem is characterized by the label ${\cal Y} = { \pm {1} }$. Specifically, given a loss function $l : \R \times {\cal Y} \rightarrow [0,\infty)$, the aim is to identify a predictor $h: {\cal X} \rightarrow \R$ with a loss
${\cal L}_{\cal D}(h) := \E_{(x,y) \sim {\cal D}} l(h(x),y)$
is small.

Moreover, when a function $h$ is defined by a parameter vector $w$,
 we denote ${\cal L}_{\cal D}(w) := {\cal L}_{\cal D}(h)$, and $l_{(x,y)}(w) := l(h(x),y)$. For a class ${\cal H}$ of predictors from $\cal X$ to $\R$, we denote 
\begin{align*} 
{\cal L}_{\cal D}({\cal H}) := \inf_{ h \in {\cal D}} {\cal L}_{\cal D} (h).
\end{align*}
\end{definition}

For classification problems, the properties of their loss function are defined as follows:

\begin{definition}[Properties of Loss Function]\label{def:properties_l}
A loss function $l$ exhibits $L$-Lipschitz characteristics if 
\begin{itemize}
    \item for all $y \in {\cal Y}$, the function $l_y(\hat{y}):= l(\hat{y},y)$ adheres to $L$-Lipschitz properties (Definition~\ref{def:L_lipschitz}).
\end{itemize}
 Similarly, it is considered convex if $l_y$ is convex (Definition~\ref{def:convex}) for each $y\in{\cal Y}$. The function is considered to have $L$-descent properties if $l_y$ is convex, $L$-Lipschitz, and twice differentiable except at a finite number of points for every $y \in {\cal Y}$.
\end{definition}

Note that \emph{shifted} ReLU activation is a \emph{descent} activation function. Furthermore, we present the definition of empirical loss on $m$ samples:

\begin{definition}[Empirical Loss]\label{def:empirical_loss}
The empirical loss for a set of $m$ points is:
\begin{align*}
    {\cal L}_S(h) :=\frac{1}{m} \sum_{i=1}^{m} l(h(x_i),y_i)
\end{align*}
Furthermore, when function $h$ is defined by a vector of parameters $w$, we denote ${\cal L}_{S}(w) := {\cal L}_{S} (h)$. For a class ${\cal H}$ of predictors $\cal X \rightarrow \R$, we denote ${\cal L}_S ({\cal H}) = \inf_{h \in {\cal H}} {\cal L}_S(h)$.
\end{definition}

In the remainder of our paper, we denote \textsc{NeuralNetworkTraining}$(\sigma, d,m,l,\eta,b,T,B)$ as the neural network training with activation $\sigma$, input dimension $d$, weight dimension $m$, loss $l$, learning rate $\eta$, SGD batch size $b$, initialization parameter $B$ and the number of iteration $T$. Additionally, this optimization process initialize weight vector as $W \sim {\cal I} (d,m,B)$(Definition~\ref{def:weight_init}).

\subsection{Kernel Spaces}\label{subsec:kernel_space}
In this section, we provide some definitions regarding kernel and kernel spaces.

\begin{definition}[Kernel]\label{def:kernel}
Let ${\cal X}$ be a given set. A \emph{kernel} is defined as a function $\k : {\cal X} \times {\cal X} \rightarrow \R$ that guarantees, for all $x_1, \ldots, x_n \in {\cal X}$, the resulting matrix $\{\k(x_i, x_j)\}_{i,j}$ is positive semi-definite. A \emph{kernel space} pertains to a Hilbert space ${\cal H}$ in which the mapping ${\cal H} \mapsto f(x)$ is bounded. The following theorem delineates a bijective correlation between kernels and kernel spaces.
\end{definition}

The succeeding theorem details a one-to-one relationship between kernels and kernel spaces.
\begin{theorem}[Kernel versus Kernel Spaces]\label{thm:biject_kernel_space}
For each kernel $\k$, a unique kernel space ${\cal H}\k$ exists such that for all $x_1, x_2 \in {\cal X}, \k(x_1, x_2) = \langle \k(\cdot, x_1), \k(\cdot, x_2) \rangle_{{\cal H}_\k}$. Similarly, for every kernel space $\cal H$, a kernel $\k$ can be found such that ${\cal H} = {\cal H}_\k$.
\end{theorem}

Within the context of ${\cal H}_\k$, the norm, and inner product are denoted by $\|\cdot\|_\k$ and $\langle \cdot, \cdot \rangle_\k$ respectively. The ensuing theorem elucidates the robust correlation between kernels and the embeddings of $X$ into Hilbert spaces.

\begin{theorem}[Kernel versus Embedding]\label{thm:biject_kernel_embed}
A function
$\k : {\cal X} \times {\cal X} \rightarrow \R$ is recognized as a kernel if and only if a mapping $\Psi: {\cal X} \rightarrow {\cal H}$ exists to some Hilbert space where
\begin{align*}
    \k(x_1, x_2) = \langle \Psi(x_1), \Psi(x_2) \rangle_{\cal H}.
\end{align*}
 In this situation, we have:
 \begin{align*}
     {\cal H}_\k = \{ f{\Psi, v} | v \in {\cal H} \} \text{, with } f_{\Psi, v}(x) = \langle v, \Psi(x) \rangle_{\cal H}.
 \end{align*}
 Furthermore, we denote $\|f\|_\k := \min \{ |v|{\cal H}: f_{\Psi, v} \}$, and the minimizer is unique.
\end{theorem}

We will leverage a certain kind of kernels which are known as \emph{inner product kernels}. These are kernels $\k: \mathbb{S}^{d-1} \times \mathbb{S}^{d-1} \rightarrow \R$ given by
$
\k(x,y) = \sum_{n=0}^{\infty} b_n \langle x,y \rangle^n
$
where $b_n > 0$ are scalars satisfying $\sum_{n=0}^{\infty} b_n < \infty$. It is well known that for any such series, $\k$ acts as a kernel. The upcoming lemma outlines a few properties of inner product kernels.

\begin{lemma}[Characteristics of Inner Product Kernel~\cite{d20}]\label{lem:property_ip_kernel}
Let $\k$ be the inner product kernel $\k(x,y) = \sum_{n=1}^{\infty} b_n \langle x,y \rangle^n$. Assuming that $b_n>0$,
\begin{itemize}
\item If $p(x) = \sum_{|\alpha| = n} a_\alpha x^\alpha$, then $p \in \{\cal H\}_\k$, and $\|p\|_\k^2 \leq \frac{1}{b_n} \sum_{|\alpha|=n} a_{\alpha}^2$.
\item For every $u \in \mathbb{S}^{d-1}$, the function $f(x) = \langle u,x \rangle^n$ resides in ${\cal H}_\k$ and $\|f\|_\k^2 = \frac{1}{b_n}$.
\end{itemize}
\end{lemma}

For a kernel $\k$ and $M>0$, we represent ${\cal H}_\k^M := \{h \in {\cal H}_\k : \|h\|_\k \leq M\}$.
Additionally, the inner product kernel space ${\cal H}_\k^M$ is a natural benchmark for learning algorithms. Then, we present the definition of Hermite polynomials and dual activation functions.

\begin{definition}[Hermite Polynomials and the Dual Activation]\label{def:hermite_dual}
Hermite polynomials $h_0, h_1, h_2, \ldots$ correspond to the sequence of orthonormal polynomials linked to the standard Gaussian measure on $\R$. Establish an activation $\sigma : \R \rightarrow \R$, we define the \emph{dual activation} of $\sigma$ as follows:
\begin{itemize}
    \item $ \hat{\sigma} (\rho) := \E_{X,Y \sim {\cal D}{\rho} }[\sigma(X) \sigma(Y)]$
where ${\cal D}{\rho}$ represents $\rho$-correlated standard Gaussian.
\item  Additionally, it stands that if $\sigma = \sum_{n=0}^{\infty} a_n h_n$, then
$
\hat{\sigma}(\rho) = \sum_{n=0}^{\infty} a_n^2 \rho^n.
$
\end{itemize}

Specifically, $\k_{\sigma}(x,y) := \hat{\sigma} (\langle x,y \rangle)$ forms an inner product kernel.
\end{definition}

\section{Related Work}
\paragraph{Sketching}
Sketching is a well-known technique to improve performance or memory complexity~\cite{cw13}. It has wide applications in linear algebra, such as linear regression and low-rank approximation\cite{cw13,nn13,mm13,rsw16,swz17,hlw17,alszz18,swz19_neurips1,swz19_neurips2,djssw19}, training over-parameterized neural network~\cite{syz21, szz21, zhasks21}, empirical risk minimization~\cite{lsz19, qszz23}, linear programming \cite{lsz19,jswz21,sy21}, distributed problems \cite{wz16,bwz16}, clustering~\cite{emz21}, generative adversarial networks~\cite{xzz18}, kernel
density estimation~\cite{qrs+22},
tensor decomposition \cite{swz19_soda}, trace estimation~\cite{jpwz21}, projected gradient descent~\cite{hmr18, xss21}, matrix sensing~\cite{dls23_sensing, qsz23},  John Ellipsoid computation \cite{ccly19,syyz22}, semi-definite programming \cite{gs22}, kernel methods~\cite{acw17, akkpvwz20, cy21, swyz21}, adversarial training~\cite{gqsw22}, cutting plane method \cite{jlsw20}, discrepany \cite{z22}, federated learning~\cite{rpuisbga20},  reinforcement learning~\cite{akl17, wzd+20,ssx21},  relational database \cite{qjs+22}.

\paragraph{Over-parameterization in Training Neural Networks.}

The investigation of the geometry and convergence patterns of various optimization methods on over-parameterized neural networks has become a significant focus within the deep learning sphere \cite{ll18,jgh18,dzps19,als19_dnn,als19_rnn,dllwz19,sy19,zczg20,os20,lsswy20, bpsw21,syz21, hlsy21, szz21,hswz22,lsy23,als+22,mosw22,z22}. The ground-breaking research by~\cite{jgh18} introduced the concept of \emph{neural tangent kernel} (NTK), a critical analytical tool in deep learning theory. By expanding the neural network's size to the extent that the network width becomes relatively large $(m\geq \Omega(n^2))$, it can be demonstrated that the training dynamic of a neural network closely mirrors that of an NTK.

\section{Proof of Correctness}\label{sec:proof_of_correctness}
At first, we present our algorithm (Algorithm~\ref{alg:nnt}) for shifted ReLU activation over two layer neural network via SGD update.

\begin{algorithm}[!ht]\caption{Neural Network Training Via Building a Data Structure of Weights 
}\label{alg:nnt}
\begin{algorithmic}[1]
\Procedure{NeuralNetworkTrainingViaPreprocessingWeights}{$d,m,B,\eta, b,T$}
\State Network parameters $d$ and $m$
\State Initialization parameter $B > 0$,
\State Learning rate $\eta > 0$, 
\State Batch size $b$, 
\State Number of steps $T > 0$
\State Access to samples from a distribution ${\cal D}$ 
\State{Sample $W \sim {\cal I}(d,m,B)$} \Comment{Definition~\ref{def:weight_init}}
\State{$b_0 \gets \sqrt{0.4 \log 2m}$} 
\State{\textsc{HalfSpaceReport hsr}} \Comment{Algorithm~\ref{alg:half_space_report}}
\State{\textsc{hsr.Init}$(\{w_r(0)\}_{r \in [2m]},2m,d)$} \Comment{This step takes ${\cal T}_{\mathsf{init}}(2m,d)$ time.}
\State
\For{$t = 1 \to T$}
    \State{Obtain a mini-batch $S_t = \{(x_i^t, y_i^t)\}_{i=1}^{b} \sim {\cal D}^b$} 
    \For{$i = 1 \to b$}
    \State{$S_{i,\text{fire}} \gets$\textsc{hsr.Query}$(x_i^t,b_0)$} \Comment{This step takes ${\cal T}_\mathsf{query} (2m,d, k_{i,t})$ time.}
    \State{$u(t)_i \gets \frac{1}{\sqrt{2m}} \sum_{r \in S_{i,\text{fire}}} a_r \cdot \sigma_{b_0}(w_r(t)^\top x_i)$}\Comment{This step takes $O(d \cdot k_{i,t})$ time}
    \EndFor
    \State{$P \gets 0^{b\times 2m}$}
    \For{$i = 1 \to b$}
        \For{$r \in S_{i,\text{fire}}$}
        \State{$P_{i,r} \gets \frac{1}{\sqrt{2m}} a_r \cdot \sigma_{b_0}' (w_r(t)^\top x_i^t)$}
        \EndFor
    \EndFor
    \State{$M \gets X \text{diag}(y-u(t))$} \Comment{$M \in \R^{d\times b}$, takes $O(bd)$ time.}
    \State{$\Delta W \gets M P$} \Comment{This step takes $O(d \cdot \nnz(P))$ time, where $\text{nnz}(P)=O(bm^{4/5})$}
    \State{$W(t+1) \gets W(t) - \eta \cdot \Delta W$.} \Comment{Backward computation.}
    \State{Let $Q \subset [2m]$, such that for each $r \in Q, \Delta W_{*,r}$ is not all zeros} \Comment{$|Q| \leq O(b m^{4/5})$}
    \For{$r \in Q$}
    \State{\textsc{hsr.Delete}$(w_r(t))$} 
    \State{\textsc{hsr.Insert}$(w_r(t+1))$} \Comment{Update the network weight.}
    \EndFor
\EndFor

\State{Choose $t \in [T]$ uniformly at random and return $W(t)$. }
\EndProcedure
\end{algorithmic}
\end{algorithm}

Next, we deliver the proof of correctness showing that Algorithm~\ref{alg:nnt} is capable of learning even polynomials of bounded norm ${\cal P}_c^M$(Definition~\ref{def:poly_bound_norm}) with nearly optimal sample complexity and network size.

\begin{theorem}[Neural Network Learning with Shifted ReLU Activation]\label{thm:shifted_relu_nnt}
Given the following conditions:
\begin{itemize}
    \item a fixed constant $c>0$ and $b_0$,
    \item $d ,M>0, R>0$, $\epsilon > 0$,
    \item the network function $2\mathsf{NN}(2m,b_0)$ with initialization as indicated in Definition~\ref{def:weight_init},
\end{itemize}
 there is a choice of
\begin{align*}
   m = \wt{O}({d^{-1} \epsilon^{-2}M^2 R^2}), T = O({\epsilon^{-2}M^2}), 
\end{align*}
 and positive values of $B$ and $\eta$. Such a selection makes sure that for every $R$-bounded distribution ${\cal D}$ (as defined in Definition~\ref{def:R_bounded_dis}) and a batch size $b$, the function $h$ obtained by Algorithm~\ref{alg:nnt} satisfies the condition that
 \begin{align*}
     \E[{\cal L}_{\cal D} (h)] \leq {\cal L}_{\cal D} ({\cal P}^M_c) + \epsilon.
 \end{align*}
\end{theorem}

From~\cite{dfs16}, for shifted ReLU activation $\sigma_{b_0}$, it holds that for every constant $c$, ${\cal P}_c^M \subset {\cal H}_{\mathrm{tk}^h_\sigma}^{O(M)}$. As a result, the following theorem implies the above theorem. 

\begin{theorem}[Neural Network Learning]\label{thm:nnt}
Given $d, M>0, R>0$ and $\epsilon > 0$, there exists a choice of 
\begin{align*}
    m = \wt{O} (\frac{M^2 R^2}{d \epsilon^2}), T = O(\frac{M^2}{\epsilon^2}),
\end{align*}
 along with $B > 0$ and $\eta>0$, such that for any batch size $b$ and $R$-bounded distribution ${\cal D}$(Definition~\ref{def:R_bounded_dis}), the function $h$ obtained by \textsc{NeuralNetworkTraining}$(\sigma, d,m,l,\eta,b,T,B)$ gives us:
\begin{align*}
\E[{\cal L}_{\cal D} (h)] \leq {\cal L}_{\cal D} ({\cal H}^{M}{\mathrm{tk}{\sigma}^{h}}) + \epsilon.
\end{align*}
\end{theorem}

We can prove this theorem by a reduction to neural tangent kernel space on the initialized weight. At first, we use $\psi_W(x)$ to denote the gradient of the function $W \mapsto h_W(x)$ with respect to the hidden weights, i.e., 
\begin{align*}
\psi_W(x) 
:=
 (u_1 \sigma'(\langle  w_1, x \rangle) x, \ldots, u_{2m} \sigma'(\langle  w_{2m}, x \rangle) x ) \in \R^{2m \times d},
\end{align*}

where we use $\sigma'(x)$ denote the first order derivative of activation $\sigma$. Moreover, we denote $f_{\psi_W, V}(x) := \langle  V, \psi_w(x) \rangle$.

Next, we show that \textsc{NeuralNetworkTraining}$(\sigma, d,m,l,\eta,b,T,B)$ is equivalent to \textsc{NeuralTangentKernelTraining}$(\sigma, d,m,l,\eta,b,T)$, with large enough initialization of the weights on neurons. We defer the proof of this lemma to Section~\ref{sec:equ_nnt_ntkt_app}.

\begin{lemma}[Equivalence for NNT and NTKT]\label{lem:eqa_nn_ntk}
If the following conditions hold
\begin{itemize}
    \item Fix a descent activation $\sigma$ as well as a convex descent loss $l$(Definition~\ref{def:convex}).
\end{itemize}
 There is a choice $B = \poly(d,m,1/\eta,T, 1/\epsilon)$, such that for every input distribution the following holds: Let $h_1,h_2$ be the functions returned by \textsc{NeuralNetworkTraining}$(\sigma, d,m,l,\eta,b,T,B)$ with parameters $d,m,\frac{\eta}{B^2}, b,B,T$ and \textsc{NeuralTangentKernelTraining}$(\sigma, d,m,l,\eta,b,T)$.
 
 Then, we have 
 \begin{itemize}
    \item $|\E[{\cal L}_{\cal D} (h_1)] - \E[{\cal L}_{\cal D} (h_2)]| < \epsilon$.
 \end{itemize}
\end{lemma}

By the above lemma, it's enough to analyze \textsc{NeuralTangentKernelTraining} 
in order to prove Theorem~\ref{thm:nnt}. To be specific, Theorem~\ref{thm:nnt} follows from the following theorem:

\begin{theorem}
Given $d, M>0,R>0$ and $\epsilon > 0$, there is a choice of $m = \wt{O} ({d^{-1}\epsilon^{-2}M^2 R^2})$, $T = O({\epsilon^{-2}M^2})$, and $\eta > 0$ which enable that for every $R$-bounded distribution $\cal D$(Definition~\ref{def:R_bounded_dis}) 
and batch size $b$, the function $h$ obtained by \textsc{NeuralTangentKernelTraining}$(\sigma, d,m,l,\eta,b,T)$ satisfies 
\begin{align*}
    \E[{\cal L}_{\cal D}(h)] \leq {\cal L}_{\cal D} ({\cal H}^M_{\mathrm{tk}^h_\sigma}) + \epsilon.
\end{align*}
\end{theorem}

In order to prove the above theorem, we prove an equivalent theorem (Theorem~\ref{thm:sgdrfs}) in the next section, where we rephrase everything in the language of the vector random feature scheme.

\subsection{Vector Random Feature Schemes}
We note that \textsc{NeuralTangentKernelTraining}$(\sigma, d,m,l,\eta,b,T)$ is SGD on top of the random embedding $\psi(W)$ that consists of $m$ i.i.d. random mappings:
\begin{align*}
\psi_W(x) = (\sigma'(\langle  W,x \rangle ) x, -\sigma' (\langle W,x \rangle) x),
\end{align*}
where $W \in \R^d$ follows a standard Gaussian distribution. For simplification, we adjust the training process to SGD on independent and identically distributed random mappings $\psi_W(x) = \sigma' (\langle W,x \rangle) x$. After the application of this mapping, inner products between different examples remain unaffected up to multiplication. As the SGD update solely depends on these inner products, analyzing the learning process within the corresponding random feature scheme framework is sufficient.
To begin with, we use random mapping $\Psi_W(x)$ to denote the random $m$-\emph{embedding} generated from $\psi$:
\begin{align*}
    \Psi_W(x) := \frac{1}{\sqrt{m}} \cdot (\psi (w_1,x), \ldots, \psi(w_m,x))
\end{align*}
where $w_1, \ldots, w_m$ are i.i.d. Next, we consider \textsc{SGDRFS}$(\psi, m,l,\eta,b,T)$ for learning the class ${\cal H}_\k$, by running SGD algorithm on these random features.
For the remainder of this section, we establish a $C$-bounded Random Feature Space (RFS) $\psi$ for a kernel $\k$ and a randomly selected $m$ embedding $\psi_w$. We adjust the notation to denote the RFS as $\psi$. The \emph{Neural Tangent Kernel (NTK)} RFS is presented by the mapping $\psi: \R^d \times \mathbb{S}^{d-1} \rightarrow \R^d$ defined by
\begin{align*} 
\psi(w,x) := \sigma' (\langle w,x \rangle)x.
\end{align*}

\begin{definition}[$m$-Kernel and $m$-Kernel Space]\label{def:m-kernel}
For a Random Feature Space (RFS) $\psi$ of a kernel $\k$, and a randomly chosen $m$ embedding $\psi_w$, the random $m$-\emph{kernel} with regard to $\psi_w$ is
\begin{align*}
\k_w(x_1, x_2) = \langle \psi_w(x_1), \psi_w(x_2) \rangle
\end{align*}
Similarly, the random $m$-\emph{kernel space} corresponding to $\psi_w$ is ${\cal H}_{\k_w}$. For every $x_1, x_2 \in {\cal X}$, we define
\begin{align*}
\k_w(x_1, x_2) = \frac{1}{m} \sum_{i=1}^{m} \langle \psi(w_i, x_1), \psi(w_i, x_2) \rangle
\end{align*}
as the average of $m$ independent random variables whose expectation is $\k(x_1, x_2)$
\end{definition}

With all the definitions, now we are ready to state the proof of correctness on the training of random feature scheme by the SGD update. Moreover, we defer the proof of Theorem~\ref{thm:sgdrfs} to Section~\ref{sec:sgdrfs_proof}.

\begin{theorem}\label{thm:sgdrfs}
Assume that
\begin{itemize}
    \item $\psi$ is a factorized (Definition~\ref{def:factorized}),
    \item $C$-bounded Random Feature Space (RFS) (Definition~\ref{def:rfs}) for $\k$,
    \item $l$ is convex (Definition~\ref{def:convex}) and $L$-Lipschitz (Definition~\ref{def:L_lipschitz}),
    \item  ${\cal D}$ has $R$-bounded marginal (Definition~\ref{def:R_bounded_dis}).
\end{itemize}
 Let $f$ be the function returned by \textsc{SGDRFS}$(\psi, m,l,\eta,b,T)$. Fix a function $f^* \in {\cal H}_k$. Then:
\begin{align*}
     \E[{\cal L}_{\cal D} (f)] 
    \leq  {\cal L}_{\cal D}(f^*) + \frac{LRC\|f^*\|_\k}{\sqrt{md}} + \frac{\|f^*\|_\k^2}{2 \eta T} + \frac{\eta L^2 C^2}{2}
\end{align*}
In particular, if $|f^|_\k \leq M$ and $\eta = \frac{M}{\sqrt{T}LC}$, we have:
\begin{align*}
     \E[{\cal L}_{\cal D} (f)] 
    \leq  L_{\cal D} (f^*) + \frac{LRCM}{\sqrt{md}} + \frac{LCM}{\sqrt{T}}
\end{align*}
\end{theorem}

\section{Proof of Running Time}\label{sec:proof_time}

In this section, we first present some definitions and properties regarding the activated neuron at each iteration. Then, we present the problem definition, algorithm, and runtime guarantee of half-space reporting. Finally, we present the runtime analysis that our algorithm has a cost per iteration sublinear in number of neurons $2m$. 

\vspace{-2mm}
\subsection{Sparsity Characterization}
\vspace{-2mm}
In this section, we show that with high probability, at every time $t \in [T]$, the number of neurons activated by shifted ReLU of each data point is sublinear in $m$. To begin with, we first present a definition which shows the set of neurons that fires at time $t$. 

\begin{definition}[Fire Set]\label{def:fire_set}
For each index $i$ within the range $[n]$, and for every timestep $t$ within the range $[T]$, we define $S_{i,\mathrm{firing}}(t)$ as a subset of $[m]$. This set corresponds to the neurons that "activate" or "fire" at time $t$. Formally, it is defined as follows:
\begin{align*}
S_{i,\mathrm{fire}}(t) := \{ r \in [m]: \langle w_r(t),x_i \rangle >b_0 \}
\end{align*}
We also define $k_{i,t}$ to be the size of the aforementioned set, i.e., $k_{i,t} := |S_{i,\mathrm{firing}}(t)|$ for all $t \in [T]$.
\end{definition}

We subsequently introduce a novel "sparsity lemma" which demonstrates that the activation function $\sigma_{b_0}$ results in the required sparsity.

\begin{lemma}[Sparsity After Initialization,~\cite{syz21}]\label{lem:sparsity_init}
Given parameter $b_0, m$, and network structure as $\mathsf{2NN}(2m,b_0)$(Definition~\ref{def:prediction_loss}), then after weight initialization, with probability 
\begin{align*} 
1-n \cdot \exp (-\Omega(m \cdot \exp(-{b_0}^2/2))),
\end{align*}
for every input $x_i$, the number of fired neurons $k_{i,0}$ is upper bounded by:
$
    k_{i,0} = O(m \cdot \exp(-b_0^2/2))
$.

\end{lemma}

Next, we present a choice of threshold $b_0$ that gives us a fired neurons set sublinear in network size $2m$:
\begin{remark}\label{remark:b_threshold}
Let $b_0 = \sqrt{0.4 \log m}$, then $k_0 = m^{4/5}$. For $t = m^{4/5}$, Lemma~\ref{lem:sparsity_init} implies that:
\begin{align*}
    \Pr[|{\cal S}_{i,\mathrm{fire}}(0)|>2m^{4/5}] \leq \exp(-\min \{mR,O(m^{4/5})\}).
\end{align*}
\end{remark}

In the forthcoming discussions, our aim is to establish that at any time instance $t \in [T]$, the count of activated neurons is sublinear with respect to $m$. Initially, we define the set of flip neurons at time $t$:

\begin{definition}[Flip Set]\label{def:flip_set}
For each index $i$ within the range $[n]$, and for each time instance $t \in [T]$, we define ${\cal S}{i,\mathrm{flip}}(t)$ as a subset of $[m]$. This set corresponds to the neurons that switch their state at time $t$. More formally, it is defined as follows:
\begin{align*}
    {\cal S}_{i,\mathrm{flip}}(t) := & ~ \{ r \in [m] : \mathrm{sgn}( \langle w_r(t), x_i \rangle - b) 
    \neq \mathrm{sign} (\langle w_r(t-1), x_i\rangle-b )\}
\end{align*}
\end{definition}

In contrast, there exist neurons that remain in the same state throughout the entire training process:

\begin{definition}[Nonflip Set]\label{def:nonflip_set}
For each index $i$ in the range $[n]$, we define $S_i$ as a subset of $[m]$. This set includes the neurons that never switch their state during the whole training process. Specifically, it is defined as follows:
\begin{align*}
    S_i & ~ :=  \{ r \in [m]: \forall t \in [T], \mathrm{sgn}(\langle w_r(t), x_i \rangle - b) \\
   & ~ =  \mathrm{sgn} (\langle w_r(0),x_i \rangle-b)\}.
\end{align*}
\end{definition}

Then, we introduce a lemma showing that the number of fired neurons $k_{i,t}$ is small for all $i \in [n], t \in[T]$ with high probability.

\begin{lemma}[Upper Bound of Fired Neuron per Iteration,~\cite{syz21}]\label{lem:bound_fired_neuron}
Let 
\begin{itemize}
    \item $b_0 \geq 0$ be a parameter,
    \item $\sigma_{b_0}(x) = \max \{x,b_0\}$ be the activation function.
\end{itemize}
 For each $i \in [n], t \in [T]$, $k_{i,t}$ is the number of activated neurons at the $t$-th iteration. For $0 < t< T$, with probability at least 
\begin{align*} 
1-n\cdot \exp(-\Omega(m) \cdot \min \{R, -\exp(-b_0^2/2)\}),
\end{align*}
$k_{i,t}$ is at most $O(m \exp(-b_0^2/2))$ for all $i\in[n]$.
\end{lemma}

\subsection{Data Structure for Half-Space Reporting}

In this section, we introduce the problem formulation, the data structure, and the time efficiency guarantees for the half-space reporting data structure. The primary objective of half-space reporting is to construct a data structure for a set $S$ in such a way that, for any given half-space $H$, the data structure can quickly identify and output the points that fall within this half-space:

\begin{definition}[Half-space Range Reporting]\label{def:half_space_report}
Given a set $S$ of $n$ points in $\R^d$, we define a half-space range reporting data structure that supports two fundamental operations:

\begin{itemize}
\item \textsc{Query}$(H)$: Provided a half-space $H \subset \R^d$, this operation returns all points within $S$ that are also contained within $H$. In other words, it outputs the intersection of $S$ and $H$.
\item \textsc{Update}: This operation pertains to modifying the set $S$ by either inserting a new point into it, or removing an existing point from it:
\begin{itemize}
\item \textsc{Insert}$(q)$: This operation inserts a point $q$ into the set $S$.
\item \textsc{Delete}$(q)$: This operation deletes the point $q$ from the set $S$.
\end{itemize}
\end{itemize}

Moreover, we denote ${\cal T}_\mathsf{init}$, ${\cal T}_\mathsf{query}$, and ${\cal T}_\mathsf{update}$ as the pre-processing time, per round query time, and per round update time for the data structure.

\end{definition}

Next, we present the formal data structure for half-space reporting. 

\begin{algorithm}[!ht]\caption{Half Space Report Data Structure}\label{alg:half_space_report}
\begin{algorithmic}[1]
\State{\textbf{data structure:} \textsc{HalfSpaceReport}}
\State{\hspace{4mm}\textbf{procedures:}}
\State{\hspace{8mm}\textsc{Init}$(S,n,d)$} \Comment{Initialize the data structure with a set $S$ of $n$ points in $\R^d$}
\State{\hspace{8mm}\textsc{Query}$(a,b)$} \Comment{$a,b \in \R^d$. Output the set $\{x \in S: \mathrm{sgn} (\langle a,x \rangle-b) \geq 0\}$}
\State{\hspace{8mm}\textsc{Add}$(x)$} \Comment{Add point $x \in \R^d$ to $S$}
\State{\hspace{8mm}\textsc{Delete}$(x)$} \Comment{Delete point $x \in \R^d$ from $S$}
\State{\textbf {end data structure}}
\end{algorithmic}

\end{algorithm}

From ~\cite{aem92}, this problem can be solved with sublinear time complexity:

\begin{corollary}[\cite{aem92}]\label{cor:half_space_query}
Given a set of $n$ points in $\R^d$, the half-space reporting problem can be solved with:
\begin{itemize}
\item A query time denoted by ${\cal T}{\mathsf{query}}(n,d,k) = O_d(n^{1-1/{\lfloor d/2 \rfloor}}+k)$, where $k = |S\cap H|$ is the number of points in the intersection of the set $S$ and half-space $H$.
\item An amortized update time denoted by ${\cal T}\mathsf{update} = O_d(\log^2 (n))$.
\end{itemize}
\end{corollary}

\subsection{Cost per iteration}

In this section, we analyze the time complexity per iteration of Algorithm~\ref{alg:nnt}.

\begin{lemma}[Running time of Algorithm~\ref{alg:nnt}]\label{lem:runtime}
Given the following:
\begin{itemize}
    \item Sample access to distribution ${\cal D} \in \R^d$,
    \item Running stochastic gradient descent algorithm (Algorithm~\ref{alg:nnt}) on $2\mathsf{NN}(2m,b_0=\sqrt{0.4 \log (2m)})$ (Definition~\ref{def:prediction_loss}) with batch size $b$, 
\end{itemize}
 then the expected cost per-iteration of this algorithm is
\begin{align*}
    \wt{O}(m^{1-\Theta(1/d)} b d)
\end{align*}
\end{lemma}

We delay the proof of Lemma~\ref{lem:runtime} to Section~\ref{sec:proof_running_time_app}.

\section{Conclusion}\label{sec:conclusion}

Deep learning is widely employed in many domains, but its model training procedure often takes an unnecessarily large amount of computational resources and time. In this paper, we design an efficient neural network training method with SGD update that has a provable convergence guarantee. By leveraging the static half-space report data structure into the optimization process of a fully connected two-layer neural network with neural tangent kernel for shifted ReLU activation, our algorithm supports sublinear time activate neuron identification via geometric search. In addition, we prove that our algorithm can converge in $O(M^2/\epsilon^2)$ time with network size quadratic in the coefficient norm upper bound $M$ and error term $\epsilon$.  As far as we are aware, our work does not have negative societal impacts. One limitation of our work is that we can study other activation functions beyond shifted ReLU in the future.

\ifdefined\isarxiv
\bibliographystyle{alpha}
\bibliography{ref}
\else
\bibliographystyle{plainnat}
\bibliography{ref}

\fi

\newpage
\onecolumn
\appendix
\section*{Appendix}
\paragraph{Roadmap.} We present some probabilistic tools in Section~\ref{sec:tools_app}. We present more preliminaries in Section~\ref{sec:prelim_app}. We present the missing proofs in Section~\ref{sec:missing_proof}.

\section{Probabilistic Inequalities}\label{sec:tools_app}

Here we present the Hoeffding bound that characterize the probability that the sum of independent random \emph{bounded} variables deviates from its true mean by a certain amount. 

\begin{lemma}[Hoeffding bound (\cite{h63})]\label{lem:hoeffding}
Let $X_1, \cdots, X_n$ denote $n$ independent bounded variables in $[a_i,b_i]$. Let $X= \sum_{i=1}^n X_i$, then we have
\begin{align*}
\Pr[ | X - \E[X] | \geq t ] \leq 2\exp \left( - \frac{2t^2}{ \sum_{i=1}^n (b_i - a_i)^2 } \right).
\end{align*}
\end{lemma}

We present Jensen's Inequality that relates the function value of a convex function on a convex combination of inputs and the value of the same convex combination of the function values on these inputs. 

\begin{lemma}[Jensen's Inequality]\label{lem:jensen}
Let $f$ be a convex function, and $\alpha \in \R^n$, such that $\sum_{i=1}^{n} \alpha_i = 1, \alpha_i \in [0,1]$. Then for all $x_1, \ldots, x_n$, we have:
\begin{align*}
    f(\sum_{i=1}^{n} \alpha_i x_i ) \leq \sum_{i=1}^{n} \alpha_i f(x_i)
\end{align*}
\end{lemma}

\section{More preliminaries}\label{sec:prelim_app}

In Section~\ref{subsec:ntk}, we introduce the necessary preliminaries related to neural tangent kernel training.
In Section~\ref{sec:prelim_app:random_feature}, we discuss some fundamental aspects of the random feature scheme.

\subsection{The Neural Tangent Kernel Training}\label{subsec:ntk}

Given fixed network parameters $\sigma, d, m$, and $B$, the \emph{neural tangent kernel} (as defined in \cite{jgh18}) for weights $W \in \R^{2m \times d}$ is defined as follows:
\begin{align*}
    \mathrm{tk}_W(x,y) := \frac{1}{2m B^2} \cdot \langle \nabla_W h_W(x), \nabla_W h_W(y) \rangle
\end{align*}
The Neural Tangent Kernel space, denoted as ${\cal H}_{\mathrm{tk}_W}$, is a linear approximation of the updates in the function value $h_W$ as a result of small alterations in the network weights $W$.
More formally, $h \in {\cal H}_{\mathrm{tk}_W}$ iff there is an $U$ such that:
\begin{align}\label{eq:h_ntk}
    h(x) = \lim_{\epsilon \rightarrow 0} \frac{1}{\epsilon} \cdot ( h_{W+\epsilon U}(x) - h_W(x) ) , ~~~~ \forall x \in \mathbb{S}^{d-1} 
\end{align}
Note that $\sqrt{m} B \cdot \|h\|_{\mathrm{tk}_W}$ is the minimal Euclidean norm of $U$ that satisfies Eq.~(\ref{eq:h_ntk}). For simplicity, we use $\mathrm{tk}_{\sigma, B}(x,y)$ to denote $\mathrm{tk}_{\sigma, d,m,B} (x,y)$, which is the \emph{expected initial neural tangent kernel}: 
\begin{align*}
    \mathrm{tk}_{\sigma, d,m,B} (x,y) := \E_{W \sim {\cal I} (d,m,B)} [\mathrm{tk}_W(x,y)]
\end{align*}

In the rest of the paper, we will abbreviate \textsc{NeuralTangentKernelTraining}$(\sigma, d,m,l,\eta,b,T)$ as neural tangent kernel training with parameters: activation function $\sigma$, input dimension $d$, weight dimension $m$, loss function $l$, learning rate $\eta$, SGD batch size $b$, and iteration number $T$. Moreover, in this optimization process, we will initialize the weight vector following $W \sim {\cal I} (d,m,1)$ and set the initial kernel weight as $V^1 = 0 \in \R^{2m \times d}$.

\subsection{Random Feature Scheme}\label{sec:prelim_app:random_feature}
In this section, we present some essential backgrounds on random feature scheme. We begin with the definition of random feature scheme with respect to kernel. 

\begin{definition}[Random Feature Scheme]\label{def:rfs}
Let $\cal X$ be a measurable space and let $\k: {\cal X} \times {\cal X} \rightarrow \R$ be a kernel. A \emph{random features scheme}(RFS) for $\k$ is a pair $(\psi, \mu)$ where $\mu$ is a probability measure on a measurable space $\Omega$, and $\psi : \Omega \times {\cal X} \rightarrow \R^d$ is a measurable function, such that:
\begin{align}\label{eq:kernel_k}
    \k(x_1, x_2) = \E_{w \sim \mu}[\langle \psi(w,x_1), \psi(w,x_2) \rangle],~\forall x_1, x_2 \in {\cal X}.
\end{align}
where $\mu$ is the standard Gaussian measure on $\R^d$, which is an RFS for the kernel $\mathrm{tk}_\sigma^h$.
\end{definition}

Next, we present the definition of $C$-bounded RFS. For activation function $\sigma$, the NTK RFS is $C$-bounded for $C = \|\sigma'\|_\infty$.
\begin{definition}[$C$-bounded RFS]\label{def:c-bounded}
We say that $\psi$ is $C$-\emph{bounded} if $\|\psi\| \leq C$. 
\end{definition}

Furthermore, we present the definition of factorized RFS. Additionally, the NTK RFS can be factorized.
\begin{definition}[Factorized RFS]\label{def:factorized}
 We say that an RFS $\psi : \Omega \times \mathbb{S}^{d-1} \rightarrow \R^d$ is \emph{factorized} if there exists a function $\psi_1 : \Omega \times \mathbb{S}^{d-1} \rightarrow \R$ such that $\psi(w,x) = \psi_1(w,x)x$. 
\end{definition}

For the remainder of this paper, we use \textsc{SGDRFS}$(\psi, m,l,\eta,b,T)$ as a shorthand to denote the Stochastic Gradient Descent in the Random Feature Space. This encapsulates the following parameters: the Random Feature Space (RFS) $\psi$, input dimension $d$, the count of random features $m$, the loss function $l$, learning rate $\eta$, SGD batch size $b$, and the number of iterations $T$. Additionally, the optimization process will initialize the weight vector as $v^1 = 0 \in \R^{m \times d}$.

\subsection{Several Instances of \texorpdfstring{$\psi$}{}}

Suppose we consider the following neural network function $f: \R^d \rightarrow \R$
\begin{align*}
f(x) = \sum_{r=1}^m a_r \sigma( \langle w_r , x \rangle ).
\end{align*}
For the ReLU activation function $\sigma$ (see \cite{dzps19,sy19}), we have 
\begin{align*}
\sigma(z) := \max\{z,0\}
\end{align*}
and
\begin{align*}
\psi(w,x) 
= & ~ x \cdot \sigma(z)' |_{z = \langle w, x\rangle } \\
= & ~ x \cdot {\bf 1}_{ \langle w, x \rangle \geq 0 }
\end{align*}

Due to recent trending of Large language models, there are a number of work study the exponential or softmax based objective function \cite{as23,bsz23,lsz23,dls23,lsx+23,ssz23,wyw+23,gsy23,zsz+23,dms23,gsx23}. Thus, we can also consider exponential activation function $\sigma$ (see \cite{gms23})
\begin{align*}
\sigma(z) := \exp(z), 
\end{align*}
and
\begin{align*}
\psi(w,x) = & ~ x \cdot \sigma(z)'|_{z = \langle w, x \rangle} \\
= & ~ x \cdot \exp(\langle w , x \rangle) 
\end{align*}

\section{Missing Proofs}\label{sec:missing_proof}

In Section~\ref{sec:equ_nnt_ntkt_app} we present the proof of the Equivalence for NNT and NTKT. In Section~\ref{sec:sgdrfs_proof}, we present the proof of Theorem~\ref{thm:sgdrfs}. In Section~\ref{sec:proof_running_time_app}, we present the proof of running time for our algorithm.

\subsection{Proof of Equivalence for NNT and NTKT}\label{sec:equ_nnt_ntkt_app}
\begin{lemma}[Equivalence for NNT and NTKT, restatement of Lemma~\ref{lem:eqa_nn_ntk}]\label{lem:eqa_nn_ntk_app}
If the following conditions hold
\begin{itemize}
    \item Fix a descent activation $\sigma$ as well as a convex descent loss $l$(Definition~\ref{def:convex}).
\end{itemize}
 There is a choice $B = \poly(d,m,1/\eta,T, 1/\epsilon)$, such that for every input distribution the following holds: Let $h_1,h_2$ be the functions returned by \textsc{NeuralNetworkTraining}$(\sigma, d,m,l,\eta,b,T,B)$ with parameters $d,m,\frac{\eta}{B^2}, b,B,T$ and \textsc{NeuralTangentKernelTraining}$(\sigma, d,m,l,\eta,b,T)$.
 
 Then, we have 
 \begin{itemize}
    \item $|\E[{\cal L}_{\cal D} (h_1)] - \E[{\cal L}_{\cal D} (h_2)]| < \epsilon$.
 \end{itemize}
\end{lemma}

\begin{proof}
For simplicity, we first prove under the assumption that the activation function $\sigma$ is twice differentiable and satisfied $\|\sigma'\|_\infty, \|\sigma''\|_\infty < M$. Then, at the end of this proof, we will show how this implies the proof for the case where the activation function is $M$-descent.

We analyze two different implementations of the \textsc{NeuralNetworkTraining}$(\cdot)$ algorithm:
The first implementation initiates with weights $W_1 = (W, u)$ sampled from distribution ${\cal I}(d,m,1)$ and adopts learning rate $\eta_1$.
The second implementation utilizes the exact same mini-batches and hyperparameters as the first one, with the exception that the output weights are scaled by a factor $B$ and the learning rate is divided by $B^2$, i.e., $W_2 = (W, Bu)$ and $\eta_2 = \eta_1/B^2$. Essentially, this transforms the network function from $h_W(x)$ to $\wt{h}_W(x) := B h_W(x)$.

Next, we show that the second implementation of \textsc{NeuralNetworkTraining} approximates \textsc{NeuralTangentKernelTraining}$(\sigma, d,m,l,\eta,b,T)$. Compared to the first implementation, the gradient of the hidden layer becomes $B$ times larger, while the gradient of the output layer remains unchanged, i.e.,
\begin{align*}
    \nabla_{W} h_{W_2} (x) = B \cdot \nabla_{W} h_{W_1} (x)\\
    \nabla_u h_{W_2} (x) = \nabla_u h_{W_1} (x)
\end{align*}
Consequently, the overall shift is scaled down by a factor of $1/B$. Thus, the optimization process operates within a sphere of radius $R/B$ around $W$, where $R$ is a polynomial in $M, d, m, 1/\eta, T, 1/\epsilon$.

Next, we examine the first-order approximation of $\wt{h}_{W}$ around the initial weight, specifically,
\begin{align*}
    |\wt{h}_{W+V} (x) - B h_w(x) - B \langle \nabla_W h_W (x) , V \rangle| \leq &~  \frac{H}{2}\|V\|^2 \\
    |\wt{h}_{W+V} (x)-B \langle \nabla_W h_W (x) , V \rangle| \leq&~\frac{H}{2} \|V\|^2
\end{align*}

The first step is derived from $h_W(x) = 0$ for the initial weight $W$ and $H$ signifies a uniform bound on the Hessian of $h_w(x)$, which is obtained from the fact that $\|\sigma'\|_\infty, \|\sigma''\|_\infty < M$. Since $R$ doesn't depend on $B$, for a sufficiently large $B$, the quadratic part $\|V\|^2 \gets 0$. Thus, we only need to consider the scenario where the optimization is conducted over the linear function $B\langle \nabla_W h_W (x) , V \rangle$ with a learning rate of $\eta/B^2$ and starting at 0. This is equivalent to \textsc{NeuralTangentKernelTraining}$(\sigma, d,m,l,\eta,b,T)$ that optimizes over the linear function $\langle \nabla_W h(W,x) ,V \rangle$ with a learning rate of $\eta$ and starting at 0.

It's important to note that any $M$-descent activation function locally ensures $\|\sigma'\|_\infty, \|\sigma''\|_\infty < M$. Additionally, if $B$ is sufficiently large, the output of the hidden layer before the activation remains largely stable throughout the entire optimization process. Given this, we don't transition into different regions that comply with $\|\sigma'\|_\infty, \|\sigma''\|_\infty < M$ for every sample in the mini-batches.

\end{proof}

\subsection{Proof of Theorem~\ref{thm:sgdrfs}}\label{sec:sgdrfs_proof}
In this section, we first present the correctness theorem for SGDRFS and its general proof by Lemma~\ref{lem:d_ran_to_d_ran}. Then, we present some definitions and lemmas to prove Lemma~\ref{lem:d_ran_to_d_ran}. Finally, we present the proof of Lemma~\ref{lem:d_ran_to_d_ran}.

\begin{theorem}[Restatement of Theorem~\ref{thm:sgdrfs}]\label{thm:sgdrfs_app} 
Assume that
\begin{itemize}
    \item $\psi$ is a factorized (Definition~\ref{def:factorized}),
    \item $C$-bounded Random Feature Space (RFS) (Definition~\ref{def:rfs}) for $\k$,
    \item $l$ is convex (Definition~\ref{def:convex}) and $L$-Lipschitz (Definition~\ref{def:L_lipschitz}),
    \item  ${\cal D}$ has $R$-bounded marginal (Definition~\ref{def:R_bounded_dis}).
\end{itemize}
 Let $f$ be the function returned by \textsc{SGDRFS}$(\psi, m,l,\eta,b,T)$. Fix a function $f^* \in {\cal H}_k$. Then:
\begin{align*}
     \E[{\cal L}_{\cal D} (f)] 
    \leq  {\cal L}_{\cal D}(f^*) + \frac{LRC\|f^*\|_\k}{\sqrt{md}} + \frac{\|f^*\|_\k^2}{2 \eta T} + \frac{\eta L^2 C^2}{2}
\end{align*}
In particular, if $|f^|_\k \leq M$ and $\eta = \frac{M}{\sqrt{T}LC}$, we have:
\begin{align*}
     \E[{\cal L}_{\cal D} (f)] 
    \leq  L_{\cal D} (f^*) + \frac{LRCM}{\sqrt{md}} + \frac{LCM}{\sqrt{T}}
\end{align*}
\end{theorem}

\begin{proof}

At first, by Hoeffding's bound~(Lemma~\ref{lem:hoeffding}), 
we have: For any $m \geq 2C^4 \epsilon^{-2} \log ({2/\delta})$, 
for every $x_1, x_2 \in {\cal X}$, we have:
\begin{align}\label{eq:kernel_apprx}
    \Pr[|\k_w(x_1, x_2) - \k(x_1, x_2)| \geq \epsilon] \leq \delta 
\end{align}

Next, we will explore how to approximate functions in ${\cal H}_\k$ using functions from ${\cal H}_{\k_w}$. For this purpose, we consider the following embedding:
\begin{align}\label{eq:psi_embed}
    x \mapsto \Psi^x | \Psi^x := \psi(\cdot, x ) \in L^2 (\Omega, \R^d) 
\end{align}
From Equation \eqref{eq:kernel_k}, we have that for any $x_1, x_2 \in {\cal X}$, $\k(x_1, x_2) = \langle \Psi^{x_1}, \Psi^{x_2} \rangle_{L^2(\Omega)}$. In particular, according to Theorem~\ref{thm:biject_kernel_embed}, for every $f \in {\cal H}_\k$, there exists a unique function $\wt{f} \in L^2 (\Omega, \R^d)$ such that:
\begin{align}\label{eq:f_embed} 
    \|\wt{f}\|_{L^2(\Omega)} = \|f\|_\k 
\end{align}

and for every $x \in {\cal X}$,
\begin{align}\label{eq:E_tilde_f}
    f(x) = \langle  \wt{f}, \Psi^x \rangle_{L^2(\Omega, \R^d)} = \E_{w \sim \mu}[ \langle \wt{f}(w), \psi(w,x) \rangle]. 
\end{align}
Then, we denote $v^*  := \frac{1}{\sqrt{m}} (\wt{f}^*(w_1), \ldots, \wt{f}^*(w_m))\in \R^{dm}$. Then, by standard results on SGD (e.g.~\cite{sb14}), we have that given $w$, 
\begin{align*}
    {\cal L}_{\cal D} (f) \leq {\cal L}_{\cal D} (f^*_w) + \frac{1}{2\eta T} \|v^*\|^2 + \frac{\eta L^2 C^2}{2}
\end{align*}
Applying the expectation over the selection of $w$, and employing Lemma~\ref{lem:d_ran_to_d_ran} along with Eq.~(\ref{eq:f_embed}), we obtain:
\begin{align*}
    {\cal L}_{\cal D}(f) \leq {\cal L}_{\cal D} (f^*) + \frac{LRC \|f^*\|_\k}{\sqrt{md}} + \frac{\|f^*\|_\k^2}{2\eta T} + \frac{\eta L^2 C^2}{2}
\end{align*}
\end{proof}

For the ease of proof, we introduce the definition of $f_w(x)$:
\begin{definition}[$f_w(x)$]\label{def:f_w_x}
Given $m,x$ and function $f$, we denote $f_w(x)$ as follows:
\begin{align*}
    f_w(x) := \frac{1}{m} \sum_{i=1}^{m} \langle \wt{f}(w_i), \psi(w_i,x) \rangle. 
\end{align*}
\end{definition}

\begin{corollary}[Function Approximation] \label{cor:fun_apprx}
For the following conditions:
\begin{itemize}
    \item for all $x \in {\cal X}$, $\E_w [|f(x) - f_w(x)|^2] \leq \frac{C^2 \|f\|_\k^2}{m}$,
    \item if ${\cal D}$ represents a distribution on $\cal X$,
\end{itemize}
 we establish that:
\begin{align*}
    \E_w [\|f-f_w\|_{2,{\cal D}}] \leq \frac{C \|f\|_\k}{\sqrt{m}} 
\end{align*}

\end{corollary}

\begin{proof}
From Equation~(\ref{eq:E_tilde_f}), we find that $\E_w [f_w (x)] = f(x)$. Additionally, for every $x$, the variance of $f_w(x)$ can be computed as follows:
\begin{align*}
    \frac{1}{m} \E_{w \sim \mu} [|\langle \wt{f}(w), \psi(w,x)\rangle |^2]  \leq & ~ \frac{C^2}{m} \E_{w \sim \mu}[|\wt{f}(w)|^2]\\
     = & ~ \frac{C^2 }{m} \|f\|^2_\k
\end{align*}
The initial step is derived from the fact that $\psi$ is $C$-bounded (see Definition~\ref{def:c-bounded}), while the concluding step is drawn from Eq.~(\ref{eq:f_embed}). Consequently, it directly leads us to:
\begin{align}\label{eq:f_variance}
    \E_w[|f(x) - f_w(x)|^2] \leq \frac{C^2}{m} \cdot \|f\|_\k^2 .
\end{align}

Additionally, when ${\cal D}$ represents a distribution on the set $\cal X$, we can derive that:
\begin{align*}
    \E_w [\|f-f_w\|_{2, {\cal D}}] \leq &~ \sqrt{\E_w [\|f-f_w\|_{2,{\cal D}}^2]}\\
    =&~ \sqrt{\E_w [\E_{x \sim {\cal D}} [|f(x) - f_w(x)|^2]]} \\
    =&~ \sqrt{\E_x [\E_w [|f(x) - f_w(x)|^2]] }\\
    \leq &~ \frac{C \|f\|_\k}{\sqrt{m}}
\end{align*}
where the first step follows from Jensen's inequality(Lemma~\ref{lem:jensen}), the second step follows from plugging in the definition of $\|\cdot\|_{2, {\cal D}}$
, the third step follows from exchanging the order of expectation, and the last step follows from using Eq.~(\ref{eq:f_variance}).
\end{proof}

Thus, $O(\frac{\|f\|_\k^2}{\epsilon^2})$ random features are adequate to ensure an expected $L^2$ distance of no more than $\epsilon$. Next, we present a situation where a $d$-dimensional random feature is as effective as $d$ one-dimensional random features. Specifically, $O(\frac{\|f\|_\k^2}{d \epsilon^2})$ random features are sufficient to guarantee an expected $L^2$ distance of at most $\epsilon$.

\begin{lemma}[Closeness of $f$ and $f_w$]\label{lem:d_ran_to_d_ran}
Assume that
\begin{itemize}
    \item $\psi : \Omega \times \mathbb{S}^{d-1}$ is factorized (Definition~\ref{def:factorized}),
    \item ${\cal D}$ is $R$-bounded distribution (Definition~\ref{def:R_bounded_dis}).
\end{itemize}
Then,
\begin{align*}
    \E_w [\|f-f_w\|_{2,{\cal D}}] \leq \sqrt{\E_w [\|f-f_w\|_{2,{\cal D}}^2]} \leq \frac{RC}{\sqrt{md}} \cdot \|f\|_\k.
\end{align*}
Additionally, if $l : \mathbb{S}^{d-1} \times Y \rightarrow [0,\infty)$ is an $L$-Lipschitz loss (as per Definition~\ref{def:L_lipschitz}), and if ${\cal D}_1$ is a distribution over $\mathbb{S}^{d-1} \times Y$ with an $R$-bounded marginal (according to Definition~\ref{def:R_bounded_dis})
then:
\begin{align*}
    \E_w [{\cal L}_{{\cal D}_1} (f_w)] \leq {\cal L}_{{\cal D}_1}(f) + \frac{LRC }{\sqrt{md}} \cdot \|f\|_\k
\end{align*}
\end{lemma}

\begin{proof}
Let us have $x \sim {\cal D}$ and $w \sim \mu$. We have: 
\begin{align*}
    \E_w [\|f-f_w\|_{2, {\cal D}}]^2 \leq &~ \E_w [\|f-f_w\|_{2,{\cal D}}^2] \\
    = &~ \E_w [\E_x [|f(x) - f_w(x)|^2] ]\\
    = &~ \E_x [\E_w [|f(x) - f_w(x)|^2]] \\
    = &~ \frac{1}{m} \cdot \E_x [\E_{w \sim \mu} [|\langle \wt{f}(w), \psi(w,x) \rangle-f(x) |^2]] \\
    \leq &~  \frac{ 1 }{m} \cdot \E_x [\E_{w \sim \mu}[|\langle \wt{f}(w), \psi(w,x) \rangle|^2]] \\
    = &~ \frac{1}{m} \cdot \E_{w \sim \mu} [\E_x [| \wt{f}(w), \psi_1 (w,x) x |^2]] \\
    \leq &~  \frac{ C^2 }{m } \cdot \E_{w\sim \mu}[ \E_x [|\langle \wt{f}(w),x \rangle|^2]] \\
    \leq &~ \frac{ C^2 R^2 } { md }  \E_{w\sim \mu} [\|\wt{f}(w)\|^2] \\
    = &~ \frac{C^2 R^2 }{md} \cdot \|f\|_\k^2
\end{align*}

where the first step comes from Jensen's inequality (Lemma~\ref{lem:jensen}
), the second step is because of plugging in the definition of $\|\cdot\|_{2, {\cal D}}$
, the third step derives from changing the order of expectations, the fourth step follows from Theorem~\ref{thm:biject_kernel_embed} and Eq.~\eqref{eq:E_tilde_f} 
, the fifth step is due to the fact that the variance is bounded by squared $L^2$-norm, the sixth step follows from that the considered RFS is factorized(Definition~\ref{def:factorized}), the seventh step is because that $\psi$ and $\psi_1$ are $C$-bounded (Definition~\ref{def:c-bounded}
), the eighth step comes from ${\cal D}$ is $R$-bounded (Definition~\ref{def:R_bounded_dis} 
), and the final step is due to Eq.~(\ref{eq:f_embed}). Taking the square root of both sides yields to:
\begin{align*}
    \E_w [\|f-f_w\|_{2, {\cal D}}] \leq \frac{CR\|f\|_\k}{\sqrt{md}}.
\end{align*}

Finally, for $L$-Lipschitz $l$(Definition~\ref{def:L_lipschitz}), and $(x,y) \sim {\cal D}_1$, then: 
\begin{align*}
    \E_w[{\cal L}_{{\cal D}_1}(f_w)] =&~ \E_w[ \E_{x,y} [l(f_w) ,y]]\\
    \leq &~ \E_w [\E_{x,y} [l(f(x),y)]] + L \E_w[ \E_x[|f(x) - f_w(x)|]]\\
    = &~ \E_{x,y} [l(f(x),y) ] + L \E_w [\E_x [ |f(x) - f_w(x) |]]\\
    = &~ {\cal L}_{{\cal D}_1} (f) + L \E_w[ \E_x [|f(x) - f_w(x)|]]\\
    \leq &~ {\cal L}_{{\cal D}_1}(f) + L \E_w[ \sqrt{\E_x[|f(x) - f_w(x)|^2]} ]\\
    \leq &~ {\cal L}_{{\cal D}_1}(f) + \frac{LCR\|f\|_\k}{\sqrt{md}}
\end{align*}
where the first step is due to definition of $\cal L$(Definition~\ref{def:binary_classification})
, the second step comes from the $L$-Lipschitzness of function $l$ (Definition~\ref{def:properties_l})

, the third step is because that $l$ is no longer a function of $w$, the fourth step follows from the definition of ${\cal L}_{{\cal D}_1}$(Definition~\ref{def:binary_classification}) 
, the fifth step follows from the fact that $\ell_1$ distance is upper bounded by $\ell_2$ distance of $f(x)$ and $f_w(x)$, the sixth step follows from Corollary~\ref{cor:fun_apprx}.
\end{proof}

\subsection{Proof of Running Time}\label{sec:proof_running_time_app}

\begin{lemma}[Running time of Algorithm~\ref{alg:nnt}, restatement of Lemma~\ref{lem:runtime}]\label{lem:runtime_app}
Given the following:
\begin{itemize}
    \item Sample access to distribution ${\cal D} \in \R^d$,
    \item Running stochastic gradient descent algorithm (Algorithm~\ref{alg:nnt}) on $2\mathsf{NN}(2m,b_0=\sqrt{0.4 \log (2m)})$ (Definition~\ref{def:prediction_loss}) with batch size $b$, 
\end{itemize}
 then the expected cost per-iteration of this algorithm is
\begin{align*}
    \wt{O}(m^{1-\Theta(1/d)} b d)
\end{align*}
\end{lemma}

\begin{proof}
The per-time complexity can be decomposed as follows:
\begin{itemize}
    \item Querying the active neuron set for $x_i \in S_t$ takes $\wt{O} (m^{1-\Theta(1/d)}bd)$ time.
    \begin{align*}
        \sum_{i=1}^{b} {\cal T}_\mathsf{query}(2m,d,k_{i,t}(S_t)) =&~b\wt{O}(m^{1-\Theta(1/d)}d)\\
        =&~ \wt{O} (m^{1-\Theta(1/d)}bd)
    \end{align*}
    where the first step follows from Corollary~\ref{cor:half_space_query}, and the final step follows from calculation. 
    \item Forward computation takes $O(b d m^{4/5})$ time.
    \begin{align*}
        \sum_{i \in [b]} O(d \cdot k_{i,t}) =&~  O(b d m^{4/5})
    \end{align*}
    where the last step is due to Lemma~\ref{lem:bound_fired_neuron}
    \item Backward computation takes $m^{4/5}bd$ time.
    \begin{itemize}
        \item Computing $M$ takes $O(bd)$ time .
        \item Computing gradient $\Delta W$ and updating $W(t+1)$ takes $O(m^{4/5}bd)$ time.
        \begin{align*}
            O(d \cdot \mathrm{nnz}(P)) = O(d\cdot b m^{4/5}) 
        \end{align*}
        where the last step is because of Lemma~\ref{lem:bound_fired_neuron}.
    \end{itemize}
    \item Updating the weight vectors in \textsc{HalfSpaceReport} data structure takes $O(bm^{4/5} \log^2 (2m))$ time. 
    \begin{align*}
        {\cal T}_\mathsf{update} \cdot (|S_{i,\mathrm{fire}}(t)|+|S_{i,\mathrm{fire}}(t+1)|) =&~ O (\log^2 (2m)) \cdot (\sum_{i \in [b]} k_{i,t} + k_{i,t+1})\\
        =&~O (\log^2 (2m)) \cdot O(b m^{4/5})\\
        =&~O(bm^{4/5} \log^2 (2m))
    \end{align*}
\end{itemize}
Summing over all the above terms gives us the per iteration running time $\wt{O}(m^{1-\Theta(1/d)} b d)$.
\end{proof}




\end{document}